\pdfoutput=1

\documentclass[11pt]{article}
\usepackage[table]{xcolor}

\usepackage[preprint]{acl}

\usepackage{times}
\usepackage{latexsym}

\usepackage[T1]{fontenc}

\usepackage[utf8]{inputenc}

\usepackage{microtype}

\usepackage{inconsolata}

\usepackage{graphicx}

\usepackage{fancyvrb, fvextra}
\usepackage{dsfont}
\usepackage{amsmath}
\usepackage{multirow}
\usepackage{algorithm}
\usepackage{algpseudocode}

\title{Improved Evidence Extraction and Metrics for \\ Document Inconsistency Detection with LLMs}

\author{
 \textbf{Nelvin Tan\textsuperscript{1}},
 \textbf{Yaowen Zhang\textsuperscript{1}},
 \textbf{James Asikin Cheung\textsuperscript{1}},
 \textbf{Fusheng Liu\textsuperscript{1}},
 \textbf{Yu-Ching Shih\textsuperscript{1}},
 \textbf{Dong Yang\textsuperscript{1}}
\\
\\
 \textsuperscript{1}American Express
 \\
 \tiny{
   \textbf{Correspondence:} \href{thongcainelvin.tan@aexp.com}{thongcainelvin.tan@aexp.com}
 }
}

\begin{document}

\maketitle

\begin{abstract}
Large language models (LLMs) are becoming useful in many domains due to their impressive abilities that arise from large training datasets and large model sizes. However, research on LLM-based approaches to document inconsistency detection is relatively limited. We address this gap by investigating evidence extraction capabilties of LLMs for document inconsistency detection. To this end, we introduce new comprehensive evidence-extraction metrics and a redact-and-retry framework with constrained filtering that substantially improves evidence extraction performance over other prompting methods. We support our approach with strong experimental results and release a new semi-synthetic dataset for evaluating evidence extraction.
\end{abstract}

\section{Introduction}

Large language models (LLMs) are becoming useful in many domains due to their impressive abilities that arise from large training datasets \cite{Li2023b} and large model sizes \cite{Naveed2023}. Despite their wide applications, research on LLM-based approaches to document inconsistency detection, a field with applications in various domains \cite{Li2018,Deubser2023}, is limited \cite{Li2024b}. Unlike natural language understanding \cite{Harabagiu2006}, where detecting inconsistencies (or contradictions) is often defined as determining the relation between a hypothesis and a piece of premise, we study the detection of inconsistencies that occur within the confines of a single input document. Furthermore, research from the field of psychology \cite{Graesser1993,Otero1992} indicates that humans struggle to identify and detect inconsistencies (or contradictions) in unfamiliar and informative texts, especially when contradictions are widely separated in long documents. This motivates a need for an automated system to tackle this challenge.

\paragraph{Problem setup.} The objective of document inconsistency detection is defined as follows: Given an input document $\boldsymbol{x}$, the goal is to correctly classify the presence of inconsistency $y\in\{\text{Yes},\text{No}\}$ and identify the set of inconsistent sentences $\mathcal{E}=\{s_1,\dots,s_k\}$ if there exists any inconsistency. Therefore, the LLM inference approach is required to output two things: (i) Classification $\hat{y}\in\{\text{Yes},\text{No}\}$ of whether there is any inconsistency. (ii) If $\hat{y}=\text{Yes}$, then the evidence set of sentences $\widehat{\mathcal{E}}=\{\hat{s}_1,\dots,\hat{s}_{k'}\}$ is also required. Note that $k'$ may not equal $k$. While improving the performances of both (i) and  (ii) are important, in this paper, we focus solely on (ii).

\paragraph{Main contributions.} While the metrics for classification are well-studied, we argue that the same is not true for evidence extraction. Therefore, we ask the following research question: \textit{How can we evaluate the evidence-extraction ability of LLMs for document inconsistency detection?} We propose new comprehensive metrics for evaluating evidence extraction, and we also propose a novel approach that substantially improves performance over other prompting methods in these metrics. To evaluate our methods and metrics in a more diverse setting, we create a semi-synthetic dataset ContraDocPaired, constructed from ContraDoc \citet{Li2024b}, that has datapoints (documents) with evidence sets of size two -- addressing the limitation of previously known dataset \cite{Li2024b} that only contains documents with an evidence set of size one.

\section{Related Work}

\paragraph{Sentence-level detection.} Most of the previous work in inconsistency detection focused on the sentence level. Specifically, prior work studied contradictions under the natural language inference (NLI) framework of evaluating contradictory pairs of sentences \cite{Dagan2005,Bowman2015}. 

\paragraph{Document-level detection.} Some NLI research has more recently been extended to document-level reasoning \cite{Yin2021,Schuster2022,Mathur2022}. However, these works do not consider inconsistency detection via LLMs holistically at the document level. The most relevant work is that of \citet{Li2024b}, which provided a dataset for document inconsistency detection and tested LLM direct prompting (DP) on it with the evidence hit metric (see \eqref{eq:EH}). However, they did not test other prompting strategies such as self-consistency and self-criticism, and their evidence hit metric alone is not comprehensive enough for evaluating evidence extraction. We fill these gaps in this paper.


\section{Metrics for Evidence Extraction} \label{sec:metrics}

\paragraph{Motivation.} Previous metrics for document inconsistency mainly account for classification performance but not evidence-extraction performance -- classification metrics are accuracy, precision, recall/true positive rate (TPR), F1 score, false positive rate (FPR), true negative rate (TNR), false negative rate and (FNR); full formulas for those are presented in Appendix \ref{app:full_exp_results}. We argue that evaluating the evidence-extraction ability of a method crucial for a holistic assessment of document inconsistency detection. For example, a method that performs perfectly (e.g., 100\% accuracy) in classification but fails to extract any relevant evidence is not helpful to the user. Worse still, this could be confusing to the user since they will always be unable to verify the source of the inconsistency, further doubting the system's conclusion.

\paragraph{Notation.} We define $\mathcal{D}=\{1,\dots,n\}$ to be the indices of the datapoints, $\mathcal{D}_+=\{i\,|\,i\in\mathcal{D}\text{ and }y_i=\text{Yes}\}$, and $\mathcal{D}_{++}=\{i\,|\,i\in\mathcal{D}\text{ and }y_i=\hat{y}_i=\text{Yes}\}$. We now define our metrics.

\paragraph{Evidence hit.} For index $i\in\mathcal{D}$, given document $\boldsymbol{x}_i$, true classification $y_i$, and true evidence set $\mathcal{E}_i$, we have the evidence hit $\text{EH}_i=\mathds{1}\{\mathcal{E}_i\subseteq \widehat{\mathcal{E}}_i\}$. We define e vidence hit rate (EHR) and evidence hit rate when correct (EHRC) as follows:
\begin{align}
    \text{EHR}&=\frac{\sum_{i\in\mathcal{D}_+}\text{EH}_i}{|\mathcal{D}_+|};\,
    \text{EHRC}&=\frac{\sum_{i\in\mathcal{D}_{++}}\text{EH}_i}{|\mathcal{D}_{++}|}. \label{eq:EH}
\end{align}
These metrics were previously introduced by \citet{Li2024b}. We argue that this metric alone is insufficient for evaluating evidence extraction. Consider an extreme case, where a method that outputs every single sentence in the document as evidence will have perfect EHR and EHRC, but this is not helpful to the user since they will be overwhelmed by too much evidence.

\paragraph{Evidence precision.} We define evidence precision $\text{EP}_i=|\mathcal{E}_i\cap\widehat{\mathcal{E}}_i|/|\widehat{\mathcal{E}_i}|$. We define evidence precision rate (EPR) and evidence precision rate when correct (EPRC) as follows:
\begin{align}
    \text{EPR}&=\frac{\sum_{i\in\mathcal{D}_+}\text{EP}_i}{|\mathcal{D}_+|};\,
    \text{EPRC}&=\frac{\sum_{i\in\mathcal{D}_{++}}\text{EP}_i}{|\mathcal{D}_{++}|}.
\end{align}
This metric is useful as it penalizes methods that produce large evidence set with many false positives. The blindspot of this metric is the scenario where the estimated evidence set is a small subset of the true evidence set. In this case, we will have $\text{EP}_i=1$ resulting in perfect EPR and EPRC, but the method is not helpful to the user since they will be missing out on many important evidence sentences.

\paragraph{Evidence recall.} We define evidence recall $\text{ER}_i=|\mathcal{E}_i\cap\widehat{\mathcal{E}}_i|/|\mathcal{E}_i|$. We define evidence recall rate (ERR) and evidence overlap rate when correct (ERRC) as follows:
\begin{align}
    \text{ERR}&=\frac{\sum_{i\in\mathcal{D}_+}\text{ER}_i}{|\mathcal{D}_+|};\,
    \text{ERRC}&=\frac{\sum_{i\in\mathcal{D}_{++}}\text{ER}_i}{|\mathcal{D}_{++}|}.
\end{align}
We argue that EHR and EHRC are useful metrics since EH, while being slightly similar to ER, is a much stricter metric that requires the entire evidence set to be correctly identified. In contrast, ER is a more lenient metric that allows partial evidence extraction to be rewarded. For example, whenever the estimated evidence is a subset of the true evidence, we have $\text{EH}=0$ and $\text{ER}>0$.


\paragraph{Remark.} Aside from EHR and EHRC, the rest are new metrics that we introduce in this paper. We have the identities and bounds, connecting classification and evidence-extraction metrics, as follows:
\begin{align*}
    \text{EHR}, \text{EPR}, \text{ERR}=\text{TPR}\times(\text{EHRC}, \text{EPRC}, \text{ERRC}) \\
    0\leq\text{EHR}, \text{EPR}, \text{ERR}\leq\text{EHRC}, \text{EPRC}, \text{ERRC}\leq1.
\end{align*}
We derive the first line in Appendix \ref{app:identities}. The second line follows from the fact that $\text{TPR}\leq0$.


\section{Methodology} \label{sec:method}

Previous prompting strategies that are viable for document inconsistency detection include direct prompting \cite{Li2024b}, self-consistency \cite{Wang2023}, and self-criticism \cite{Tan2025a, Li2024a}. Details of how they are adapted to document inconsistency is presented in Appendix \ref{app:prompting_strategies}. Further investigation of how temperature of self-consistency affects the performance of document inconsistency detection is also presented in Appendix \ref{app:self_con}.

Next, we explain the key components of the redact-and-retry framework: redact, retry, and filter. The idea of using redact-and-retry with a filter call is loosely inspired by \citet{Zhang2024}, where they used a sequence of LLM calls (i.e., worker agents) followed by a final LLM call (i.e., manager agent) to synthesize the outputs of prior LLM calls. We start by defining the function $\text{Redact}(\boldsymbol{x},\widehat{\mathcal{E}})$ which takes in a document $\boldsymbol{x}$ and outputs the same document, but with all the sentences present in $\widehat{\mathcal{E}}$ removed. 

\paragraph{Redact-and-retry.} Our algorithm is shown in Algorithm \ref{alg:RnR}, and a visual representation is presented in Figure \ref{fig:RnR} -- output from lines 3 and 7 can be from \textit{any} prompting strategy, but for simplicity, we use direct prompting since it minimizes the number of LLM calls needed; the prompt is given in Appendix \ref{app:prompts}. The intuition is that redacting preserves the sentence order in the document and also reduces the search space for the LLM, which we expect would make it easier for the LLM to detect the remaining inconsistencies (if there are any left). Since $\widehat{\mathcal{E}}_i^{(1)}\subseteq\bigcup_j\widehat{\mathcal{E}}_i^{(j)}$, EHR, EHRC, ERR, and ERRC can only improve from the output of line 3.

\paragraph{Filter.} We explore the idea of applying a filter (i.e., extra LLM call) to the evidence set output $\bigcup_j\widehat{\mathcal{E}}_i^{(j)}$ of redact-and-retry -- this is shown visually in Figure \ref{fig:RnR+F} in the appendix. The role of the filter call is for the LLM to re-analyze the sentences in $\bigcup_j\widehat{\mathcal{E}}_i^{(j)}$ and only output the sentences that it thinks are truly inconsistent. The filter call is expected to reduce the size of the evidence set, but at the cost of an additional LLM call. We study 2 types of filter calls: (i) \textit{Unconstrained.} The filter call is allowed to return any number of sentences, including zero. If it returns an empty evidence list, then we will change the initial classification from Yes to No. \textit{Constrained.} The filter call is constrained to return at least one sentence. Hence, we do not change the LLM's initial classification (from redact-and-retry). The prompts for both filter calls are given in Appendix \ref{app:prompts}. We conduct an ablation study in Appendix \ref{app:RnR_exp} and found that redact-and-retry with the constrained filter is the best all-round approach, in terms of balancing evidence extraction performance with the number of sentences extracted.

\begin{algorithm}
    \caption{Redact-and-Retry Algorithm}
    \label{alg:RnR}
    \begin{algorithmic}[1] 
        \State \textbf{Input:} document $\boldsymbol{x}_i$.
        \State Initialize $\boldsymbol{x}_i^{(1)}=\boldsymbol{x}_i$ and $j=1$.
        \State Get $\hat{y}_i^{(1)},\widehat{\mathcal{E}}_i^{(1)}=\text{LLM}(\boldsymbol{x}_i^{(1)})$.
        \While{$\hat{y}_i^{(j)}=\text{Yes}$}
            \State $j=j+1$.
            \State $\boldsymbol{x}_i^{(j)}=\text{Redact}(\boldsymbol{x}_i^{(j-1)},\widehat{\mathcal{E}}_i^{(j-1)})$.
            \State $\hat{y}_i^{(j)},\widehat{\mathcal{E}}_i^{(j)}=\text{LLM}(\boldsymbol{x}_i^{(j)})$.
        \EndWhile
        \State Set $\hat{y}_i=\hat{y}_i^{(1)}$ and $\widehat{\mathcal{E}}_i=\bigcup_j\widehat{\mathcal{E}}_i^{(j)}$.
        \State \Return $\hat{y}_i$ and $\widehat{\mathcal{E}}_i$.
    \end{algorithmic}
\end{algorithm}

\begin{figure*}[t]
    \centering
    \includegraphics[width=2.0\columnwidth]{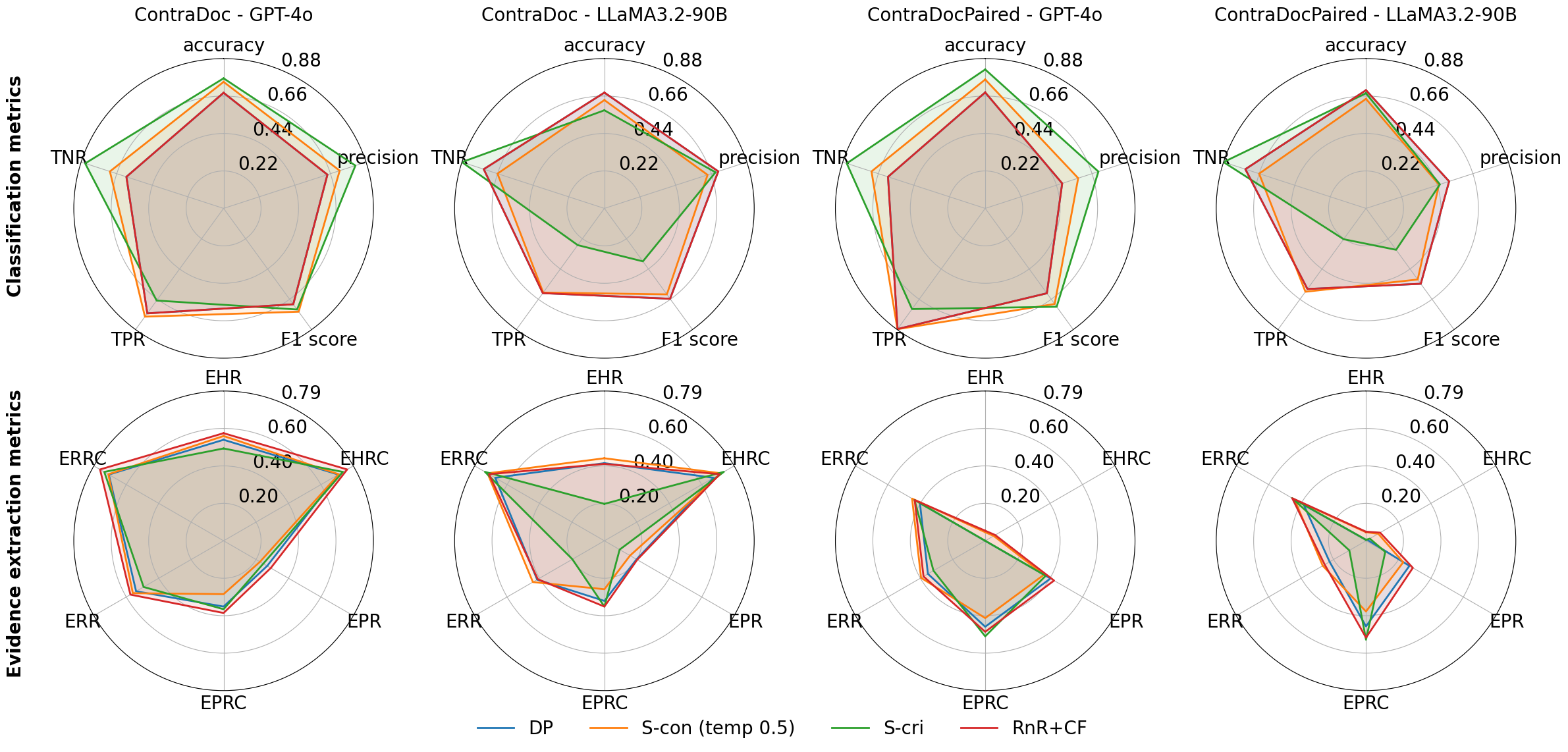}
    \caption{1st and 2nd rows corresponds to classification metrics and evidence-extraction metrics, respectively. 1st, 2nd, 3rd, and 4th columns correspond to ContraDoc with GPT-4o, ContraDoc with LLaMA3.2-90B, ContraDocPaired with GPT-4o, and ContraDocPaired with LLaMA3.2-90B, respectively.}
    \label{fig:diff_methods}
\end{figure*}

\section{Experiments} \label{sec:experiments}

Due to the non-deterministic nature of LLMs, the LLM sentence outputs might not be exactly the same as the sentence in the input documents. Therefore, we allow approximate matching instead of exact matching in certain aspects of our implementation: (i) For the computation of $\text{EH}_i$, when checking whether $\mathcal{E}_i\subseteq \widehat{\mathcal{E}}_i$, we check whether each sentence in $\mathcal{E}_i$ has a match in $\widehat{\mathcal{E}}_i$ that has a cosine similarity of at least 0.8 using the TF-IDF vectorizer \cite{Jones1972}. If it is a yes for all such sentences, then we determine that $\mathcal{E}_i\subseteq \widehat{\mathcal{E}}_i$. (ii) For the computation of $\text{EP}_i$, when checking whether $\mathcal{E}_i\cap \widehat{\mathcal{E}}_i$, we check whether each sentence in $\mathcal{E}_i$ has a match in $\widehat{\mathcal{E}}_i$ that has a cosine similarity of at least 0.8 using the TF-IDF vectorizer \cite{Jones1972}. If it is a yes, then that sentence is in $\mathcal{E}_i\cap \widehat{\mathcal{E}}_i$. (iii) For the computation of $\text{Redact}(\boldsymbol{x},\widehat{\mathcal{E}})$, when checking whether a sentence $s$ in $\boldsymbol{x}$ is in $\widehat{\mathcal{E}}$, we check whether there exists a sentence in $\widehat{\mathcal{E}}$ that has a cosine similarity of at least 0.8 with $s$ using the TF-IDF vectorizer \cite{Jones1972}.

\subsection{Dataset and Models} \label{sec:dataset_models}

We use the ContraDoc dataset from \citet{Li2024b}, which contains 449 positive (inconsistent) documents and 442 negative (consistent) documents -- every positive document has exactly one inconsistent sentence. The average number of sentences per document for the positive documents, negative documents, and all documents are 38.1, 36.8, and 37.5 respectively. We test our approach using 2 different LLMs with temperature set to 0, unless stated otherwise, for consistency of results: GPT-4o (strong) and LLaMA3.2-90B (weak). More details on ContraDoc are in Appendix \ref{app:contradoc}.

\paragraph{ContraDocPaired.} ContraDoc has $|\mathcal{E}_i|=1$ for all $i\in\mathcal{D}$, resulting in $\text{EH}_i=\text{ER}_i$ for all $i$. Hence, we are unable to highlight experimentally the differences between EHR (EHRC) and ERR (ERRC). We were unable to find such a dataset where $|\mathcal{E}_i|>1$, and therefore created a semi-synthetic dataset, ContraDocPaired\footnote{Dataset link: \url{https://anonymous.4open.science/r/ContraDocPaired-0BC4/README.md}.}, constructed from ContraDoc, that has datapoints (documents) with $|\mathcal{E}_i|=2$ for all $i\in\mathcal{D}$. Details on the creation process are in Appendix \ref{app:ContraDocPaired}.

\paragraph{Shorthands.} We use the shorthands DP for direct prompting, S-con for self consistency, S-cri for self criticism, and RnR, RnR+UF, and RnR+CF for redact-and-retry with no filter, an unconstrained filter, a constrained filter, respectively.

\subsection{Comparison of different methods}

We compare DP, S-con with temperature 0.5, S-cri, and RnR+CF. We picked RnR+CF since it is the best all-rounder performer among the RnR variants (from experiments in Appendix \ref{app:RnR_exp}). Results are displayed in Figure \ref{fig:diff_methods} -- the numbers behind the plots in Figure \ref{fig:diff_methods} are provided in Appendix \ref{app:full_exp_results}. We make some observations: (i) For Classification, there is no clear winner among the different methods. One clear observation is that for LLaMA3.2-90B, S-cri has a much lower TPR and F1 score score compared to the rest, while also having a much higher TNR. This distinction is less obvious in both datasets for GPT-4o. (ii) For evidence extraction, RnR+CF is arguably the best performer among the different methods for both datasets and both LLMs. For every radar circle on the second row, we observe that the RnR+CF line is always around the furthest line for all metrics.

We further investigate the number of sentences extracted by the different methods in Appendix \ref{app:diff_methods_sentences}, showing that RnR+CF controls the number of sentences extracted to be low, further cementing RnR+CF's viability.

\section{Conclusion}

We introduced a semi-synthetic dataset and comprehensive evidence-extraction metrics for document inconsistency detection, and showed empirically that our RnR+CF approach, applied to DP's outputs, outperforms other methods in evidence extraction. It would be interesting to investigate how the true evidence set size affects the performance of different methods, and how the performance changes with varying lengths of input documents. This would require careful creation of new datasets, which we leave to future work.

\newpage

\section{Limitations}

We used two LLMs, which might seem limited, but argue that it may be sufficient since they are of different strength levels (i.e., GPT-4o is a strong LLM, while LLaMA3.2-90B is a weak LLM). We considered two datasets that are correlated, which may limit the generalizability of our findings. While we would like to test our approach on other datasets where $|\mathcal{E}_i|>1$, we were unable to find such a viable dataset -- document inconsistency datasets are generally not readily available, as the field is relatively new compared to other LLM applications. Creating ContraDocPaired is a step towards addressing this gap, and we hope that it can be used by future research in this field.

With regards to the RnR framework, while the framework is generalizable to any prompting strategy (see lines 3 and 7 of Algorithm \ref{alg:RnR}), we tested it with DP for simplicity -- we believe this to be the most practical and viable since combining RnR with other prompting strategies would result in the number of LLM calls being too high for practical applications. Nevertheless, it is still interesting to know whether we can do better with other prompting strategies. This is left to future work.




\bibliography{custom}

\appendix

\section{Visualization of Algorithms} \label{app:visuals}

\begin{figure*}[t]
    \centering
    \includegraphics[width=2.0\columnwidth]{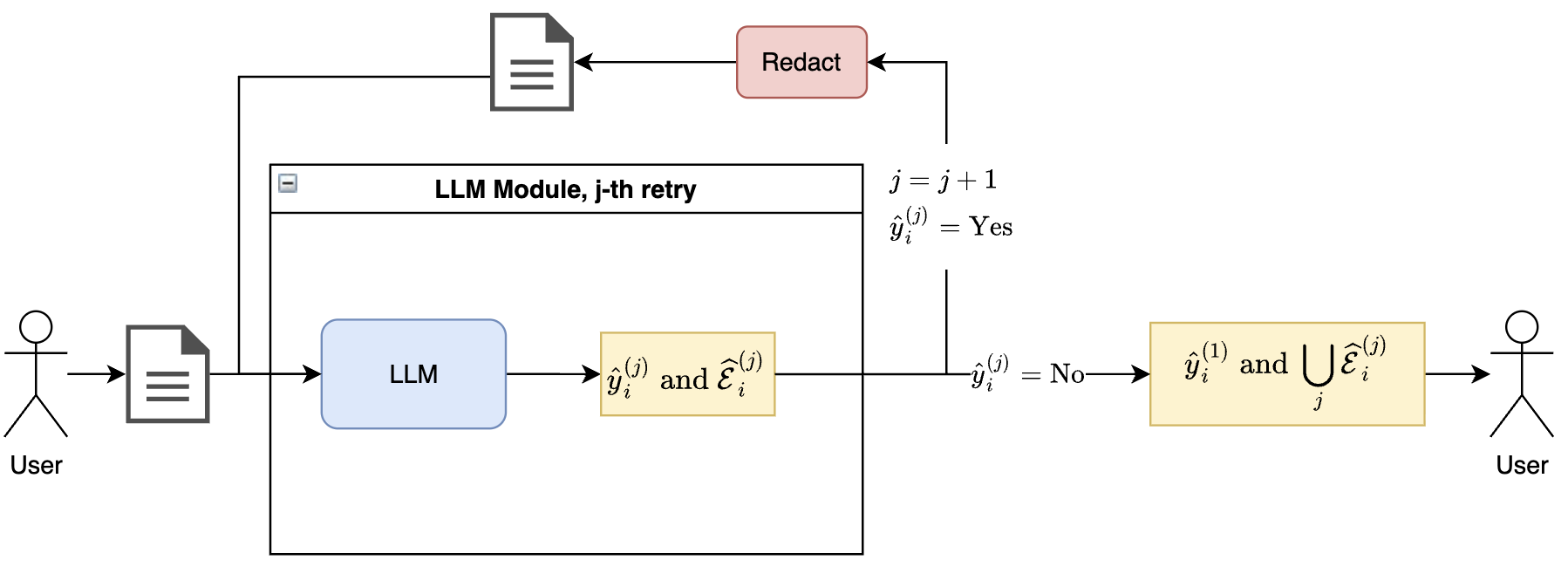}
    \caption{Algorithm \ref{alg:RnR} visualized.}
    \label{fig:RnR}
\end{figure*}

\begin{figure}[t]
    \centering
    \includegraphics[width=1.0\columnwidth]{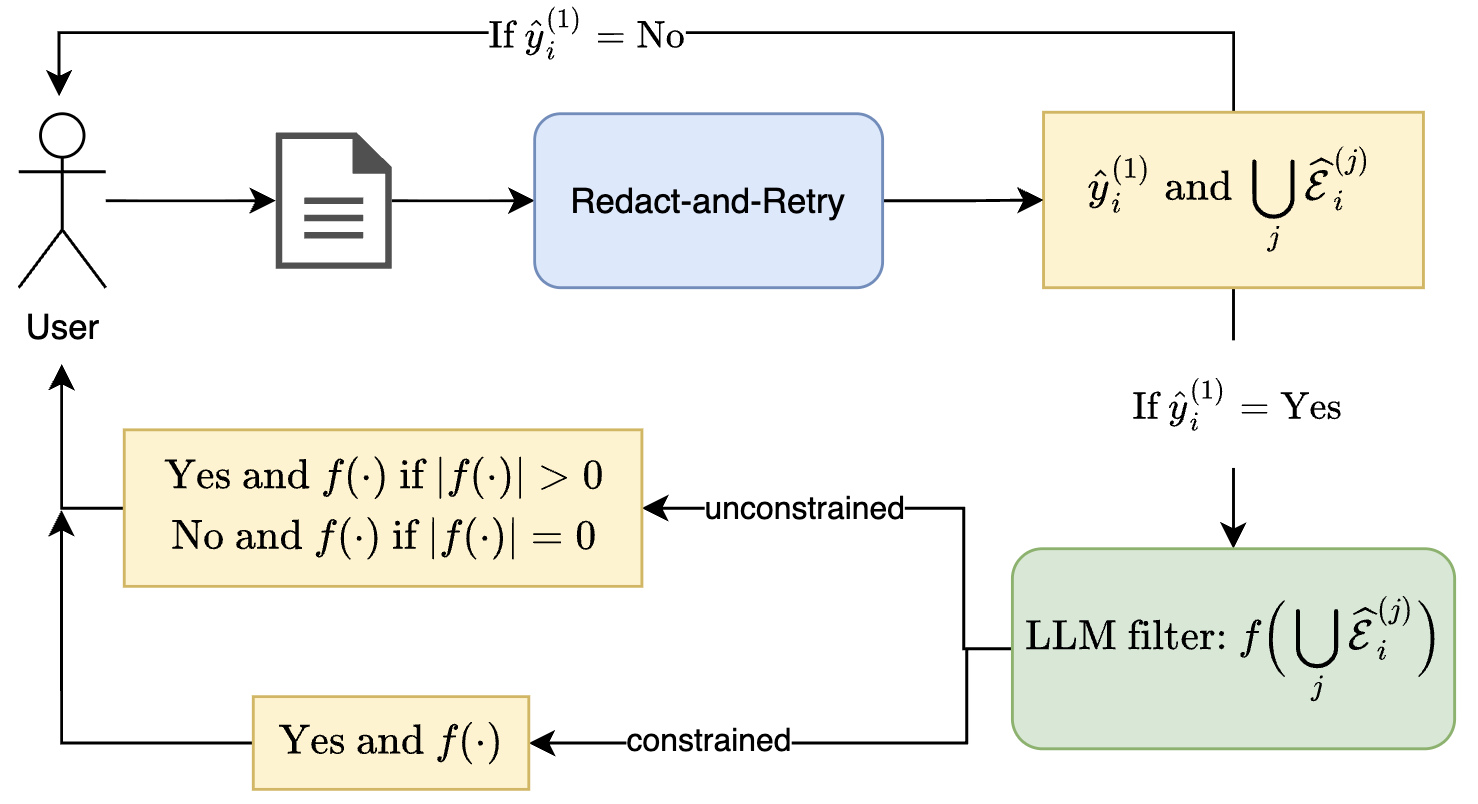}
    \caption{Algorithm \ref{alg:RnR} with LLM filter call visualized.}
    \label{fig:RnR+F}
\end{figure}

The visualizations of RnR and RnR+filter (both unconstrained and constrained filters) are provided in Figure \ref{fig:RnR} and Figure \ref{fig:RnR+F}, respectively.

\section{Derivation of Identities} \label{app:identities}

\paragraph{Identity 1.} $\text{EHR}=\text{TPR}\times\text{EHRC}$. \\
\textit{Proof:} We have
\begin{align*}
    \text{EHR}
    &\stackrel{(a)}{=}\frac{1}{|\mathcal{D}_+|}\sum_{i\in\mathcal{D}_+}\text{EH}_i \\
    &\stackrel{(b)}{=}\frac{1}{|\mathcal{D}_+|}\sum_{i\in\mathcal{D}_{++}}\text{EH}_i \\
    &\stackrel{(c)}{=}\frac{1}{\text{TP}+\text{FN}}\sum_{i\in\mathcal{D}_{++}}\text{EH}_i \\
    &=\frac{\text{TP}}{\text{TP}+\text{FN}}\cdot\frac{1}{\text{TP}}\sum_{i\in\mathcal{D}_{++}}\text{EH}_i \\
    &\stackrel{(d)}{=}\text{TPR}\times\text{EHRC},
\end{align*}
where:
\begin{itemize}
    \item (a) uses the definition of EHR.
    \item (b) uses the fact that $\text{EH}_i=0$ for all $i\in\mathcal{D}_+\setminus\mathcal{D}_{++}$ since $\widehat{\mathcal{E}}_i=\emptyset$ when $\hat{y}_i=\text{No}$.
    \item (c) uses the fact that $\text{TP}+\text{FN}=|\mathcal{D}_+|$.
    \item (d) uses the definition of TPR and EHRC, along with the fact that $\text{TP}=|\mathcal{D}_{++}|$.
\end{itemize}

\paragraph{Identity 2.} $\text{EPR}=\text{TPR}\times\text{EPRC}$. 
\textit{Proof:} This can be derived by using similar steps to the proof of identity 1, and noting that $\text{EP}_i=0$ for all $i\in\mathcal{D}_+\setminus\mathcal{D}_{++}$ since $\widehat{\mathcal{E}}_i=\emptyset$ when $\hat{y}_i=\text{No}$.

\paragraph{Identity 3.} $\text{ERR}=\text{TPR}\times\text{ERRC}$. 
\textit{Proof:} This can be derived by using similar steps to the proof of identity 1, and noting that $\text{ER}_i=0$ for all $i\in\mathcal{D}_+\setminus\mathcal{D}_{++}$ since $\widehat{\mathcal{E}}_i=\emptyset$ when $\hat{y}_i=\text{No}$.

\section{Details of Prompting Strategies} \label{app:prompting_strategies}

We adapt known prompting strategies to document inconsistency detection, and then we introduce our novel approach.
\begin{itemize}
    \item \textbf{Direct prompting \cite{Li2024b}.} We directly prompt the LLM to output the classification and evidence set. Prompt used is provided in Appendix \ref{app:prompts}.
    \item \textbf{Self-consistency \cite{Wang2023}.} We prompt the LLM multiple times (using direct prompting) and take the majority vote for classification. If the majority classification is a `yes', then we take the union of the evidence sets across all prompts as the final evidence set.
    \item \textbf{Self-criticism \cite{Tan2025a, Li2024a}.} We use the `If-or-Else' framework introduced in \citet{Li2024a} that is shown to outperform other self-criticism frameworks \cite{Tan2025a, Li2024a}. This prompting strategy first asks the model to assess its own confidence; if confident, it keeps the original answer, else it triggers self-correction and revision. This conditional gating avoids unnecessary revisions that can degrade correct answers, using confidence as a control signal for when self-correction should occur. Prompts used are provided in Appendix \ref{app:prompts}.
\end{itemize}

\section{Creation Process of ContraDocPaired} \label{app:ContraDocPaired} 

We create a new semi-synthetic dataset ContraDocPaired, constructed from ContraDoc, that has datapoints (documents) with evidence set size two. We want to create $\lfloor n/2\rfloor$ new datapoints. Each new datapoint is formed by combining two original positive datapoints in ContraDoc (so it contains two evidence sentences). We aim to keep the maximum new datapoint length as small as possible, since the text being too long might exceed the context length of our LLMs. To achieve our goal, we employ an optimal-greedy algorithm where we first sort the positive datapoints by length. Next, we pair the longest remaining item with the shortest remaining item to create a new positive datapoint in ContraDocPaired. Pairing the largest with smallest minimizes the largest pair sum (standard proof by exchange argument). 

There are 224 positive (inconsistent) documents and 442 negative (consistent) documents -- every positive document has exactly two inconsistent sentences. The average number of sentences per document for the positive documents, negative documents, and all documents are 76.0, 36.5, and 50.0 respectively.

\paragraph{Remark.} The ContraDoc dataset is license under Apache License, Version 2.0 (source: \url{https://github.com/ddhruvkr/CONTRADOC/blob/main/LICENSE}) -- a permissive free software license that allows others to freely use, modify, and redistribute the work.

\section{Effect of Temperature on Self-Consistency} \label{app:self_con}

\begin{figure*}[t]
    \centering
    \includegraphics[width=2.0\columnwidth]{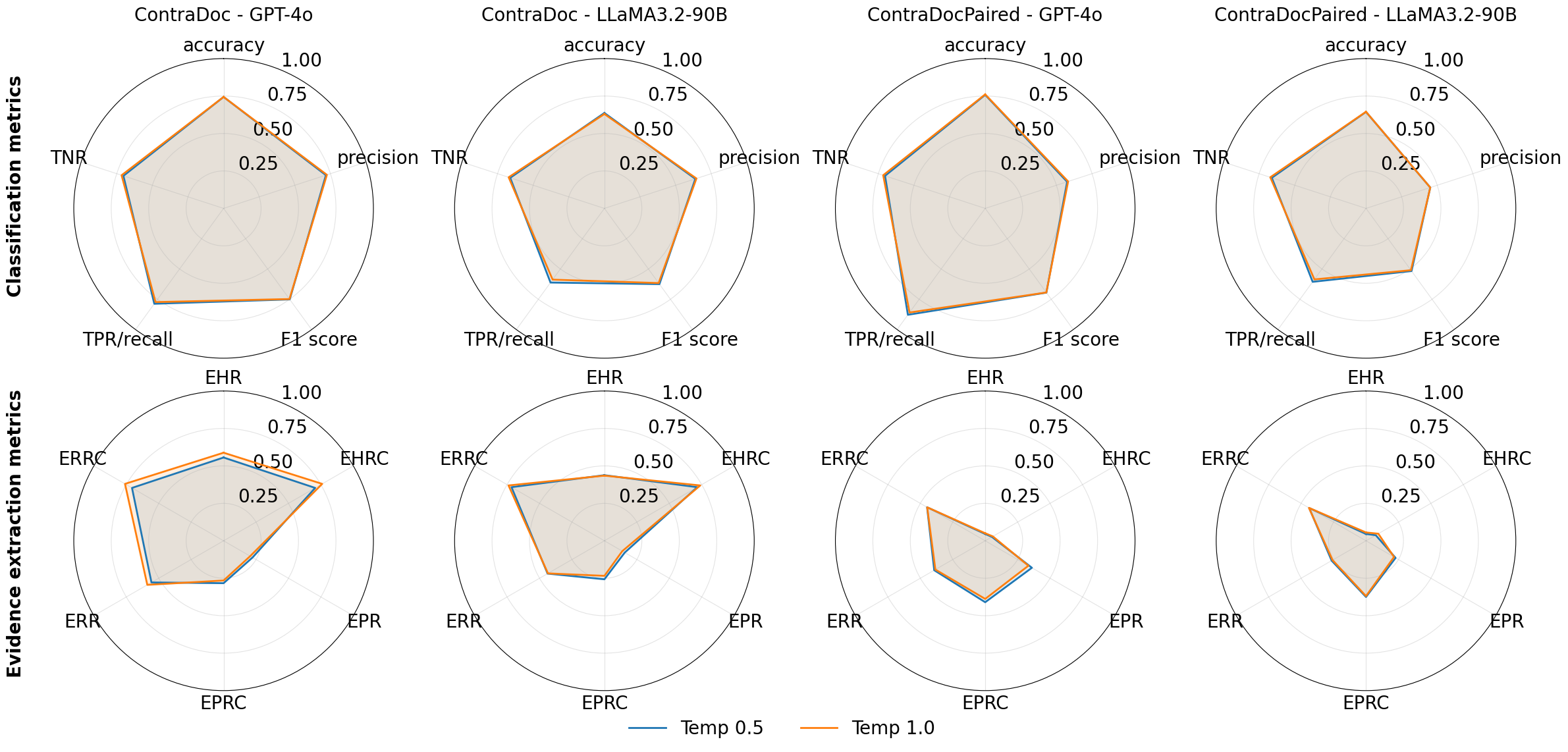}
    \caption{Radar charts for S-con with different temperatures. First row corresponds to classification metrics, second row corresponds to evidence-extraction metrics, first column corresponds to ContraDoc with GPT-4o, second column corresponds to ContraDoc with LLaMA3.2-90B, third column corresponds to ContraDocPaired with GPT-4o, and fourth column corresponds to ContraDocPaired with LLaMA3.2-90B.}
    \label{fig:self_con}
\end{figure*}

\begin{figure}[t]
    \centering
    \includegraphics[width=1.05\columnwidth]{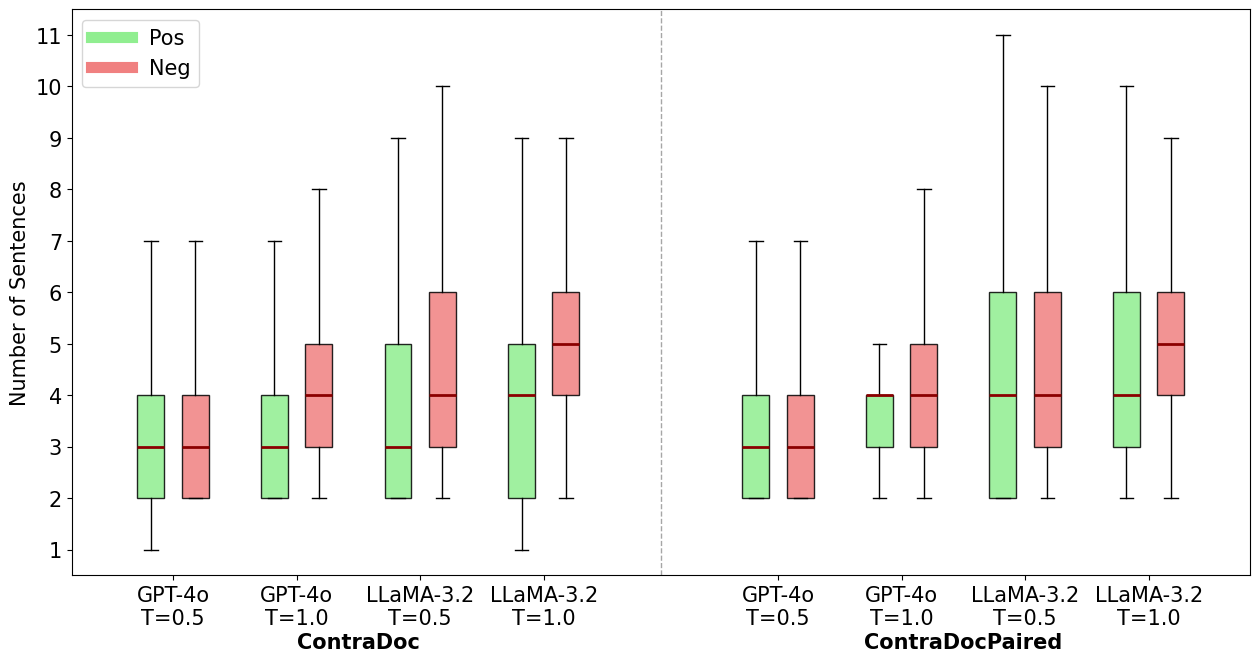}
    \caption{Boxplot of the number of sentences extracted for self-consistency with temperature 0.5 and 1.0. `Neg' refers to datapoints $i$ such that truth $y_i=\text{No}$ but prediction $\hat{y}_i=\text{No}$. `Pos' refers to datapoints $i$ such that $y_i=\hat{y}_i=\text{Yes}$.}
    \label{fig:self_con_boxplot}
\end{figure}

We run S-con with temperature 0.5 (less random) and 1.0 (more random). Results are displayed in Figure \ref{fig:self_con} -- the numbers behind the plots in Figure \ref{fig:self_con} are provided in Appendix \ref{app:full_exp_results}. We observe that for both LLMs and both datasets, the performance for both classification and evidence extraction are around the same. We further investigate the effects of temperature by studying the number of sentences extracted -- this is shown visually in Figure \ref{fig:self_con_boxplot}. We observe the following:
\begin{itemize}
    \item For positive datapoints, the number of sentences extracted stayed around the same, with the 75th percentile (top of the box) staying the same across the two temperatures for both LLMs, and the median and 25th percentile (bottom of the box) sometimes increasing with an increase in temperature from 0.5 to 1.0.
    \item For negative datapoints, the number of sentences extracted generally increased with an increase in temperature from 0.5 to 1.0 for both LLMs, with the median and 25th percentile always increasing. The 75th percentile stayed around the same for both temperatures in the case of LLaMA-3.2-90B, but increased in the case of GPT-4o.
\end{itemize}
We conclude that the additional sentences extracted when we increase the temperature from 0.5 to 1.0 are rarely helpful since the classification and evidence extraction performance stayed around the same. Therefore, it might be more advantageous to use a lower temperature with the self-consistency approach.

\section{Ablation Study for Redact-and-Retry Framework.} \label{app:RnR_exp}

\begin{figure*}[t]
    \centering
    \includegraphics[width=2.0\columnwidth]{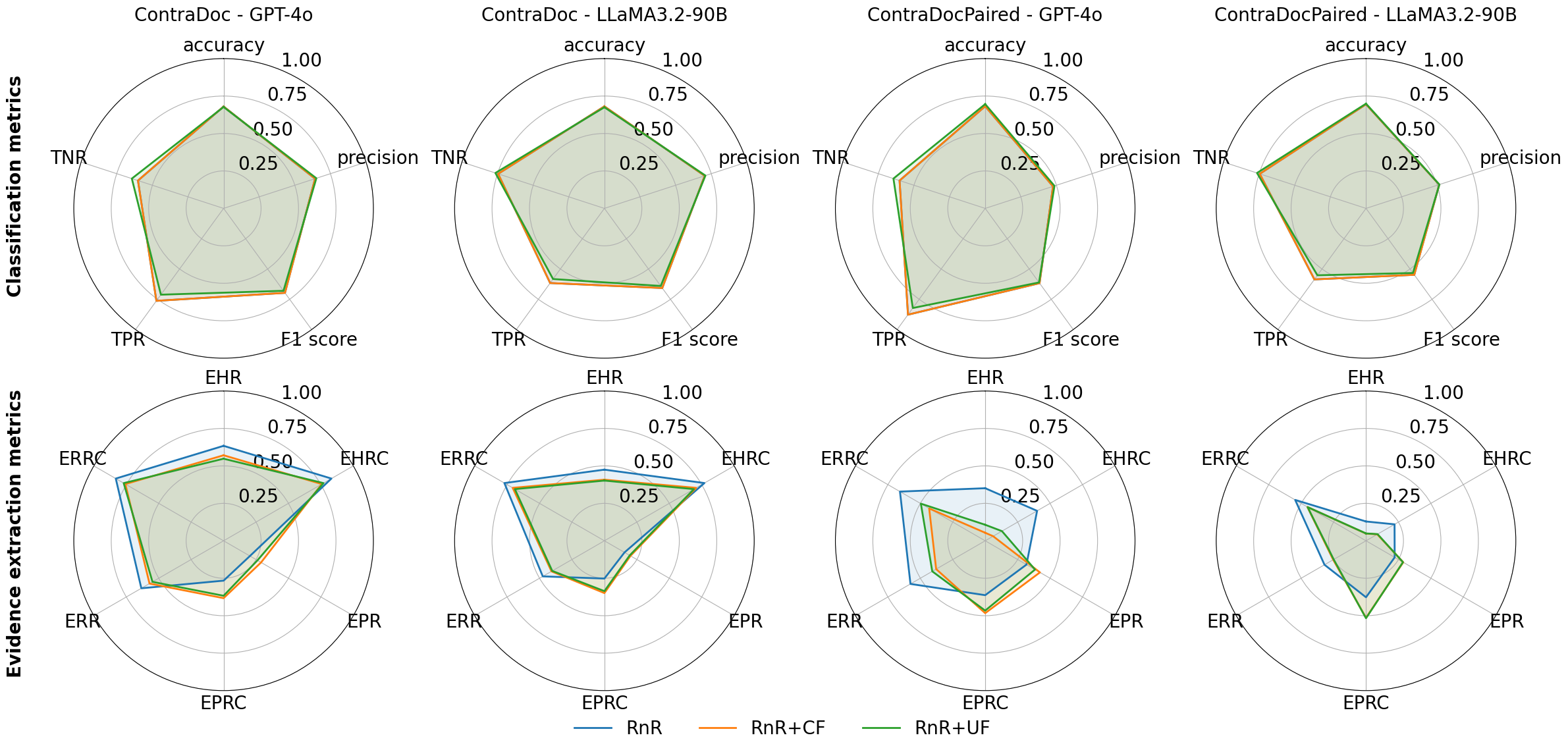}
    \caption{Radar charts for ablation study of redact-and-retry framework. First row corresponds to classification metrics, second row corresponds to evidence-extraction metrics, first column corresponds to ContraDoc with GPT-4o, second column corresponds to ContraDoc with LLaMA3.2-90B, third column corresponds to ContraDocPaired with GPT-4o, and fourth column corresponds to ContraDocPaired with LLaMA3.2-90B.}
    \label{fig:RnR_ablation}
\end{figure*}

\begin{figure}[t]
    \centering
    \includegraphics[width=1.05\columnwidth]{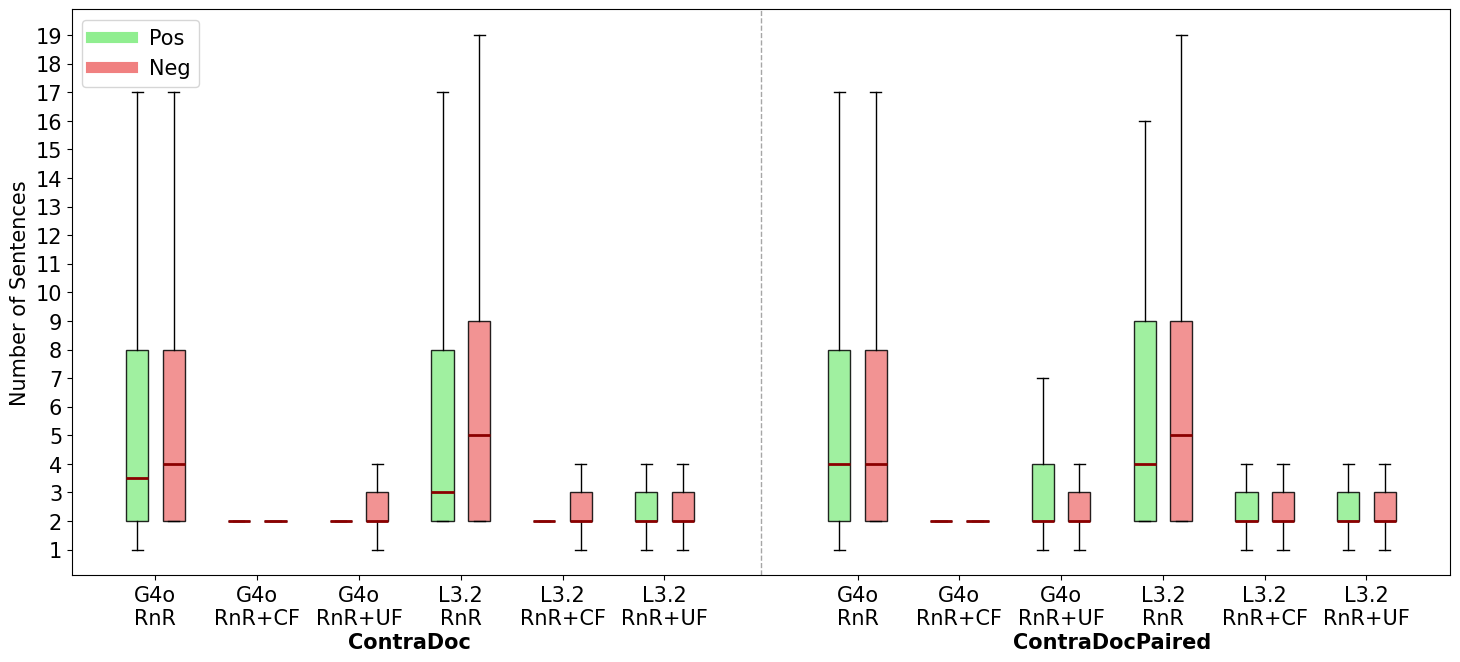}
    \caption{Boxplot of the number of sentences extracted for RnR, with and without filters. `Neg' refers to datapoints $i$ such that truth $y_i=\text{No}$ but prediction $\hat{y}_i=\text{No}$. `Pos' refers to datapoints $i$ such that $y_i=\hat{y}_i=\text{Yes}$.}
    \label{fig:RnR_ablation_boxplot}
\end{figure}

\begin{figure}[t]
    \centering
    \includegraphics[width=1.0\columnwidth]{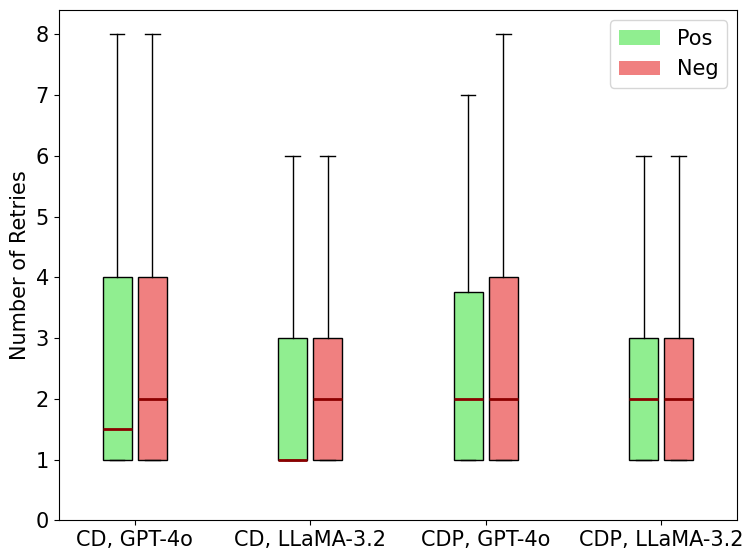}
    \caption{Boxplot for number of retries for RnR. `Pos' refers to datapoints $i$ such that truth $y_i=\text{Yes}$.`Neg' refers to datapoints $i$ such that truth $y_i=\text{No}$. CD and CDP are shorthands for ContraDoc and ContraDocPaired respectively.}
    \label{fig:num_retries}
\end{figure}

We study the effect of the filter call in the redact-and-retry framework by comparing RnR, RnR+UF, and RnR+CF. Results are displayed in Figure \ref{fig:RnR_ablation} -- the numbers behind the plots in Figure \ref{fig:RnR_ablation} are provided in Appendix \ref{app:full_exp_results}. We make some observations:
\begin{itemize}
    \item For both datasets and both LLMs, the use of the filter call does not have any significant effect on the classification metrics.
    \item For both datasets and both LLMs, the use of the filter call leads to an evidence-extraction tradeoff: a better EPR and EPRC, but a lower EHR, EHRC, ERR, and ERRC. This effect is more pronounced in ContraDocPaired than in ContraDoc for both LLMs.
    \item For ContraDoc and for both LLMs, RnR+CF has slightly better evidence extraction performance compared to RnR+UF. For ContraDocPaired and for both LLMs, both RnR+CF and RnR+UF have similar evidence extraction performance.
\end{itemize}

\paragraph{Number of sentences.} Figure \ref{fig:RnR_ablation_boxplot} summarizes the number of sentences extracted for RnR, RnR+UF, and RnR+CF. RnR+CF and RnR+UF extract way less sentences than RnR, which is expected since the filter call is designed to reduce the number of sentences. RnR+CF extracts slightly less sentences than RnR+UF, despite having similar evidence extraction performance, making CF a better filter choice. Overall, RnR+CF is the best all-rounder performer among the three approaches.

\paragraph{Number of retries.} Figure \ref{fig:num_retries} shows visually how the number of retries for RnR are distributed. We make the following observations:
\begin{itemize}
    \item The stronger the LLM, the more retries it will make. The same conclusion follows through for RnR+UF and RnR+CF since their number of retries are always just 1 more than RnR. This implies that stronger LLMs are more aggressive in detecting inconsistencies.
    \item The median number of retries for both LLMs does not exceed 3 for both datasets, implying that it is well controlled. Even though the size of the evidence set doubled from ContraDoc to ContraDocPaired, the number of retries did not increase significantly, which is a good sign for the scalability of RnR (and also RnR+UF and RnR+CF).
\end{itemize}

\paragraph{Error analysis of the filters.} To provide a detailed error analysis of the unconstrained filter (UF), we compute the following metrics: (1) Rate wrong (-) to correct (+) given flipped classification:
\begin{align*}
    R_{\text{UF}}(-\rightarrow+\,|\,\text{flip})=\frac{\#\{-\rightarrow+\}}{\text{\#flips}}.
\end{align*}
A similar definition applies to $R_{\text{CF}}(-\rightarrow+\,|\,\text{flip})$, where instead of UF, we look at the constrained filter. (2) Rate correct (+) to wrong (-) given flipped classification:
\begin{align*}
    R_{\text{UF}}(+\rightarrow-\,|\,\text{flip})=\frac{\#\{+\rightarrow-\}}{\text{\#flips}}.
\end{align*}
A similar definition applies to $R_{\text{CF}}(+\rightarrow-\,|\,\text{flip})$, where instead of UF, we look at the constrained filter. (3) Rate of UF keeping a true evidence $e$ given the earlier RnR sub-algorithm found $e$:
\begin{align*}
    R_{\text{UF}}(\text{$e$ kept}\,|\,\text{$e$ found})=\frac{\text{\#$e$ kept}}{\text{\#$e$ found}}.
\end{align*}
A similar definition applies to $R_{\text{CF}}(\text{$e$ kept}\,|\,\text{$e$ found})$, where instead of UF, we look at the constrained filter. (4) Rate of UF discarding true evidence given the earlier RnR sub-algorithm found true evidence:
\begin{align*}
    R_{\text{UF}}(\text{$e$ discarded}\,|\,\text{$e$ found})=\frac{\text{\#$e$ discarded}}{\text{\#$e$ found}}.
\end{align*}
A similar definition applies to $R_{\text{CF}}(\text{$e$ discarded}\,|\,\text{$e$ found})$, where instead of UF, we look at the constrained filter. The results are displayed in Table \ref{tab:F_error_analysis_CD} and \ref{tab:F_error_analysis_CDP}. We make the following observations: (i) From the flip rates, we observe that CF performed better for ContraDoc whereas UF performed better for ContraDocPaired. (ii) $R_{\text{UF}}(\text{$e$ kept}|\text{$e$ found})$ is better with GPT-4o vs LLaMA3.2-90B for ContraDoc (for both UF and CF), but better with LLaMA3.2-90B vs GPT-4o for ContraDocPaired (for both UF and CF).

\begin{table}[t]
    \centering
    \begin{tabular}{lcc}
    \hline
     & \textbf{G4o} & \textbf{L3.2}\\
    \hline
    $R_{\text{UF}}(-\rightarrow+|\text{flip})$ & 0.472 & 0.380 \\
    $R_{\text{UF}}(+\rightarrow-|\text{flip})$ & 0.528 & 0.619 \\
    $R_{\text{CF}}(-\rightarrow+|\text{flip})$ & 0 & 0 \\
    $R_{\text{CF}}(+\rightarrow-|\text{flip})$ & 0 & 0 \\
    $R_{\text{UF}}(\text{$e$ kept}|\text{$e$ found})$ & 0.861 & 0.847 \\
    $R_{\text{UF}}(\text{$e$ disc.}|\text{$e$ found})$ & 0.139 & 0.153 \\
    $R_{\text{CF}}(\text{$e$ kept}|\text{$e$ found})$ & 0.908 & 0.884 \\
    $R_{\text{CF}}(\text{$e$ disc.}|\text{$e$ found})$ & 0.092 & 0.116 \\
    \hline
    \end{tabular}
    \caption{Error analysis of the filters for ContraDoc. G4o and L3.2 are shorthands for GPT-4o and LLaMA3.2-90B respectively. `disc.' is the shorthand for discarded.}
    \label{tab:F_error_analysis_CD}
\end{table}

\begin{table}[t]
    \centering
    \begin{tabular}{lcc}
    \hline
     & \textbf{G4o} & \textbf{L3.2}\\
    \hline
    $R_{\text{UF}}(-\rightarrow+|\text{flip})$ & 0.654 & 0.615 \\
    $R_{\text{UF}}(+\rightarrow-|\text{flip})$ & 0.346 & 0.385 \\
    $R_{\text{CF}}(-\rightarrow+|\text{flip})$ & 0 & 0 \\
    $R_{\text{CF}}(+\rightarrow-|\text{flip})$ & 0 & 0 \\
    $R_{\text{UF}}(\text{$e$ kept}|\text{$e$ found})$ & 0.710 & 0.784 \\
    $R_{\text{UF}}(\text{$e$ disc.}|\text{$e$ found})$ & 0.290 & 0.216 \\
    $R_{\text{CF}}(\text{$e$ kept}|\text{$e$ found})$ & 0.658 & 0.790 \\
    $R_{\text{CF}}(\text{$e$ disc.}|\text{$e$ found})$ & 0.342 & 0.210 \\
    \hline
    \end{tabular}
    \caption{Error analysis of the filters for ContraDocPaired. G4o and L3.2 are shorthands for GPT-4o and LLaMA3.2-90B respectively. `disc.' is the shorthand for discarded.}
    \label{tab:F_error_analysis_CDP}
\end{table}


\section{Number of Sentences from Different Methods} \label{app:diff_methods_sentences}

\begin{figure*}[t]
    \centering
    \includegraphics[width=2.0\columnwidth]{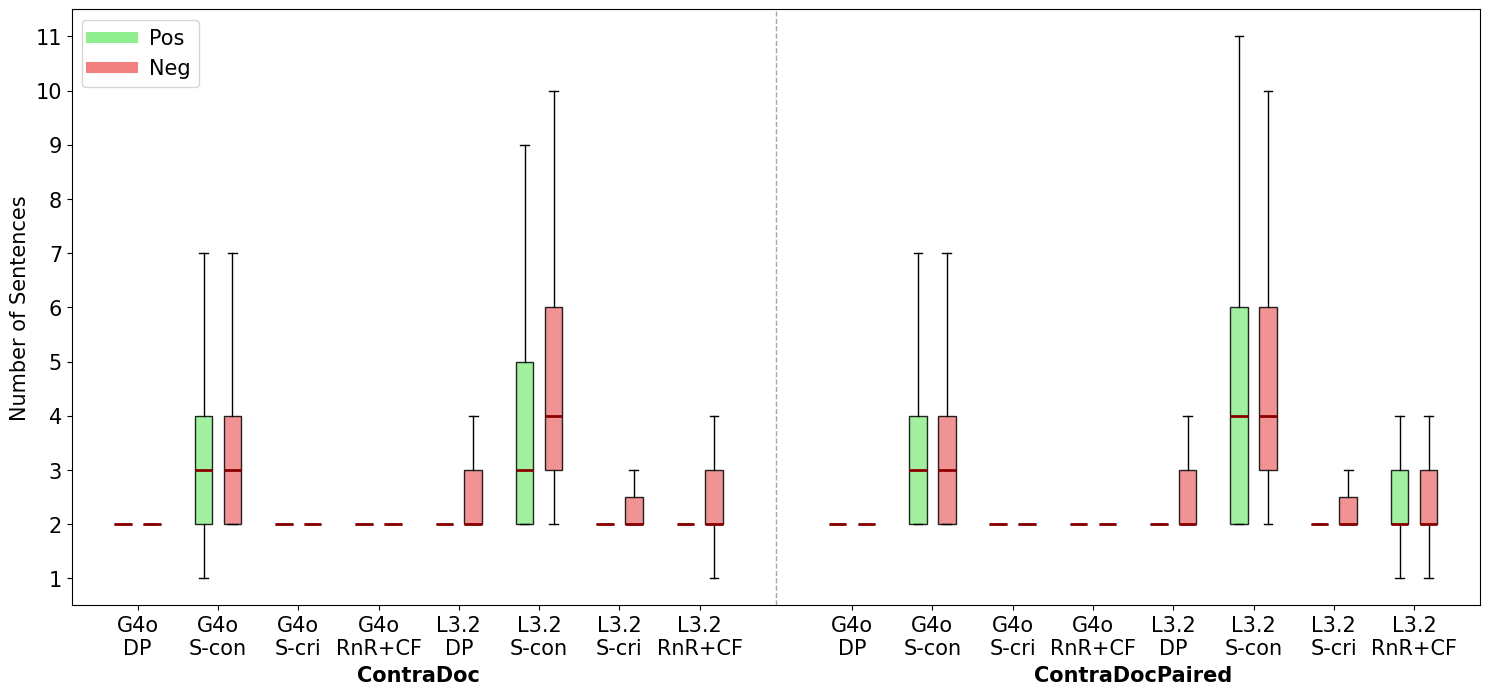}
    \caption{`Neg' refers to datapoints $i$ such that truth $y_i=\text{No}$ but prediction $\hat{y}_i=\text{No}$. `Pos' refers to datapoints $i$ such that $y_i=\hat{y}_i=\text{Yes}$.}
    \label{fig:diff_methods_boxplot}
\end{figure*}

We investigate the number of sentences extracted for the different methods -- this is shown visually in Figure \ref{fig:diff_methods_boxplot}. We observe the following: (i) There is a general trend that when LLMs (from all approaches) misclassify a negative document as positive, they tend to extract more sentences compared to when they correctly classify a positive document. (ii) Comparing across methods, S-cri extracts the least number of sentences, followed by DP, RnR+CF, and S-con -- with the gap between RnR+CF and S-con being the most significant. Since RnR+CF is not much worse than DP and S-cri in the number of sentences extracted, and coupled with its strong extraction capabilities, we argue that RnR+CF is the best all-round approach.

\section{Full Experiment Results} \label{app:full_exp_results}

Details on the classification metrics are as follows:
\begin{itemize}
    \item Accuracy $\frac{1}{|\mathcal{D}|}\sum_{i\in\mathcal{D}}\mathds{1}\{y_i=\hat{y}_i\}$, where $\mathcal{D}$ is the set of all data points.
    \item Precision $\text{TP}/(\text{TP}+\text{FP})$, where TP is the number of true positives and FP is the number of false positives.
    \item Recall/true positive rate $\text{TPR}=\text{TP}/(\text{TP}+\text{FN})$ where FN is the number of false negatives.
    \item F1 score $(2\times\text{Precision}\times\text{Recall})/(\text{Precision}+\text{Recall})$.
    \item FPR false positive rate $\text{FP}/(\text{TN}+\text{FP})$.
    \item TNR true negative rate $\text{TN}/(\text{TN}+\text{FP})$, where TN is the number of true negatives.
    \item FNR false negative rate $\text{FN}/(\text{TP}+\text{FN})$.
    \item EHR, EHRC, EPR, EPRC, and AECR are defined in Section \ref{sec:metrics}.
\end{itemize}
The full experimental results for self-consistency are presented in Table \ref{tab:selfcon_results}, the full experimental results for the ablation study of the redact-and-retry framework are presented in Table \ref{tab:ContraDoc_ablation} and Table \ref{tab:ContraDocPaired_ablation}, and the full experimental results for the comparison of different methods are presented in Table \ref{tab:ContraDoc_diff_methods} and Table \ref{tab:ContraDocPaired_diff_methods}. All results in the tables are in 3 significant figures.

\begin{table*}[t]
    \centering
    \begin{tabular}{lcccccccccccc}
    \hline
     & \multicolumn{2}{c}{CD, S-con, 0.5}
     & \multicolumn{2}{c}{CD, S-con, 1.0}
     & \multicolumn{2}{c}{CDP, S-con, 0.5}
     & \multicolumn{2}{c}{CDP, S-con, 1.0} \\
     
     & \textbf{G4o} & \textbf{L3.2}
     & \textbf{G4o} & \textbf{L3.2}
     & \textbf{G4o} & \textbf{L3.2}
     & \textbf{G4o} & \textbf{L3.2} \\
    \hline

    accuracy & 0.744 & 0.637 & 0.744 & 0.630 & 0.758 & 0.644 & 0.761 & 0.645 \\
    precision & 0.718 & 0.638 & 0.725 & 0.646 & 0.575 & 0.451 & 0.582 & 0.451 \\
    F1 score & 0.751 & 0.625 & 0.748 & 0.616 & 0.695 & 0.517 & 0.694 & 0.510 \\
    TPR/recall & 0.787 & 0.612 & 0.773 & 0.588 & 0.878 & 0.606 & 0.859 & 0.586 \\
    FPR & 0.297 & 0.339 & 0.284 & 0.328 & 0.297 & 0.339 & 0.284 & 0.328 \\
    TNR & 0.703 & 0.661 & 0.716 & 0.672 & 0.703 & 0.661 & 0.716 & 0.672 \\
    FNR & 0.215 & 0.388 & 0.227 & 0.412 & 0.122 & 0.394 & 0.141 & 0.414 \\ \hline
    EHR & 0.556 & 0.438 & 0.588 & 0.435 & 0.0457 & 0.0455 & 0.0505 & 0.0556 \\
    EHRC & 0.706 & 0.716 & 0.760 & 0.739 & 0.0520 & 0.075 & 0.0588 & 0.0948 \\
    EPR & 0.223 & 0.157 & 0.205 & 0.138 & 0.360 & 0.228 & 0.333 & 0.216 \\
    EPRC & 0.283 & 0.257 & 0.265 & 0.234 & 0.410 & 0.375 & 0.388 & 0.369 \\
    ERR &  0.556 & 0.438 & 0.588 & 0.435 & 0.393 & 0.265 & 0.384 & 0.258 \\
    ERRC & 0.706 & 0.716 & 0.760 & 0.739 & 0.448 & 0.438 & 0.447 & 0.440 \\
    avg \#sen, pos & 3.00 & 3.94 & 3.54 & 4.34 & 3.17 & 4.61 & 3.72 & 4.93 \\
    avg \#sen, neg & 3.64 & 4.85 & 4.57 & 5.26 & 3.64 & 4.85 & 4.57 & 5.26 \\
    avg \#sen, all & 3.18 & 4.27 & 3.82 & 4.67 & 3.37 & 4.74 & 4.08 & 5.11 \\
    \hline
    \end{tabular}
    \caption{\textbf{Self-Consistency.} We use the shorthand CD for ContraDoc, CDP for ContraDocPaired, S-con for self-consistency, and the numbers 0.5 and 1.0 refer to the temperature used for self-consistency. G4o and L3.2 are shorthands for GPT-4o and LLaMA3.2-90B respectively. `pos' refers to datapoints $i$ such that $y_i=\text{Yes}$, `neg' refers to datapoints $i$ such that $y_i=\text{No}$, and `all' refers to all datapoints.}
    \label{tab:selfcon_results}
\end{table*}

\begin{table*}[t]
    \centering
    \begin{tabular}{lcccccc}
    \hline
     & \multicolumn{2}{c}{RnR}
     & \multicolumn{2}{c}{RnR+CF}
     & \multicolumn{2}{c}{RnR+UF} \\
     & \textbf{G4o} & \textbf{L3.2}
     & \textbf{G4o} & \textbf{L3.2}
     & \textbf{G4o} & \textbf{L3.2} \\
    \hline

    accuracy & 0.680 & 0.681 & 0.680 & 0.681 & 0.677 & 0.675 \\
    precision & 0.642 & 0.704 & 0.642 & 0.704 & 0.651 & 0.709 \\
    F1 score & 0.697 & 0.657 & 0.697 & 0.657 & 0.680 & 0.640 \\
    TPR/recall & 0.763 & 0.616 & 0.763 & 0.616 & 0.712 & 0.583 \\
    FPR & 0.399 & 0.254 & 0.399 & 0.254 & 0.356 & 0.235 \\
    TNR & 0.601 & 0.746 & 0.601 & 0.746 & 0.644 & 0.765 \\
    FNR & 0.237 & 0.384 & 0.237 & 0.384 & 0.288 & 0.417 \\ \hline
    EHR & 0.634 & 0.475 & 0.571 & 0.408 & 0.548 & 0.402 \\
    EHRC & 0.831 & 0.771 & 0.757 & 0.707 & 0.770 & 0.690 \\
    EPR & 0.203 & 0.155 & 0.289 & 0.201 & 0.261 & 0.195 \\
    EPRC & 0.266 & 0.252 & 0.383 & 0.349 & 0.367 & 0.336 \\
    ERR & 0.634 & 0.475 & 0.571 & 0.408 & 0.548 & 0.402 \\
    ERRC & 0.831 & 0.771 & 0.757 & 0.707 & 0.770 & 0.690 \\
    avg\#sen, pos & 6.40 & 6.42 & 2.33 & 2.57 & 2.85 & 2.71 \\
    avg\#sen, neg & 6.41 & 6.85 & 2.26 & 2.67 & 2.90 & 2.84 \\
    avg\#sen, all & 6.40 & 6.55 & 2.31 & 2.60 & 2.87 & 2.75 \\
    avg\#retries, pos & 2.95 & 2.47 & 3.95 & 3.47 & 3.95 & 3.47 \\
    avg\#retries, neg & 2.88 & 2.55 & 3.88 & 3.55 & 3.88 & 3.55 \\
    avg\#retries, all & 2.92 & 2.50 & 3.92 & 3.50 & 3.92 & 3.50 \\
    \hline
    \end{tabular}
    \caption{\textbf{ContraDoc ablation.} G4o and L3.2 are shorthands for GPT-4o and LLaMA3.2-90B respectively. `pos' refers to datapoints $i$ such that $y_i=\text{Yes}$, `neg' refers to datapoints $i$ such that $y_i=\text{No}$, and `all' refers to all datapoints.}
    \label{tab:ContraDoc_ablation}
\end{table*}

\begin{table*}[t]
    \centering
    \begin{tabular}{lcccccccc}
    \hline
     & \multicolumn{2}{c}{DP}
     & \multicolumn{2}{c}{S-con (temp 0.5)}
     & \multicolumn{2}{c}{S-cri}
     & \multicolumn{2}{c}{RnR+CF} \\
     
     & \textbf{G4o} & \textbf{L3.2}
     & \textbf{G4o} & \textbf{L3.2}
     & \textbf{G4o} & \textbf{L3.2}
     & \textbf{G4o} & \textbf{L3.2} \\
    \hline

    accuracy & 0.680 & 0.681 & 0.744 & 0.637 & 0.765 & 0.577 & 0.680 & 0.681  \\
    precision & 0.642 & 0.704 & 0.718 & 0.638 & 0.815 & 0.687 & 0.642 & 0.704  \\
    F1 score & 0.697 & 0.657 & 0.751 & 0.625 & 0.735 & 0.385 & 0.697 & 0.657  \\
    TPR/recall & 0.763 & 0.616 & 0.787 & 0.612 & 0.67 & 0.267 & 0.763 & 0.616  \\
    FPR & 0.399 & 0.254 & 0.297 & 0.339 & 0.144 & 0.119 & 0.399 & 0.254  \\
    TNR & 0.601 & 0.746 & 0.703 & 0.661 & 0.856 & 0.881 & 0.601 & 0.746  \\
    FNR & 0.237 & 0.384 & 0.215 & 0.388 & 0.33 & 0.733 & 0.237 & 0.384 \\ \hline
    EHR & 0.536 & 0.412 & 0.556 & 0.438 & 0.490 & 0.196  & 0.571 & 0.408 \\
    EHRC & 0.703 & 0.669 & 0.706 & 0.716 & 0.731 & 0.732 & 0.757 & 0.707 \\
    EPR & 0.267 & 0.197 & 0.223 & 0.157 & 0.243 & 0.0934  & 0.289 & 0.201 \\
    EPRC & 0.349 & 0.319 & 0.283 & 0.257 & 0.363 & 0.349  & 0.383 & 0.349 \\
    ERR & 0.536 & 0.412 & 0.556 & 0.438 & 0.490 & 0.196  & 0.571 & 0.408 \\
    ERRC & 0.703 & 0.669 & 0.706 & 0.716 & 0.731 & 0.732 & 0.757 & 0.707 \\
    avg\#sen, pos & 2.11 & 2.33 & 3.00 & 3.94 & 2.08 & 2.36 & 2.33 & 2.57 \\
    avg\#sen, neg & 2.22 & 2.61 & 3.64 & 4.85 & 2.15 & 2.41 & 2.26 & 2.67 \\
    avg\#sen, all & 2.15 & 2.41 & 3.18 & 4.27 & 2.09 & 2.37 & 2.31 & 2.60 \\
    avg\#retries, pos & NA & NA & NA & NA & NA & NA & 3.95 & 3.47 \\
    avg\#retries, neg & NA & NA & NA & NA & NA & NA & 3.88 & 3.55 \\
    avg\#retries, all & NA & NA & NA & NA & NA & NA & 3.92 & 3.50 \\
    \hline
    \end{tabular}
    \caption{\textbf{ContraDoc.} G4o and L3.2 are shorthands for GPT-4o and LLaMA3.2-90B respectively. `pos' refers to datapoints $i$ such that $y_i=\text{Yes}$, `neg' refers to datapoints $i$ such that $y_i=\text{No}$, and `all' refers to all datapoints.}
    \label{tab:ContraDoc_diff_methods}
\end{table*}

\begin{table*}[t]
    \centering
    \begin{tabular}{lcccccc}
    \hline
     & \multicolumn{2}{c}{RnR}
     & \multicolumn{2}{c}{RnR+CF}
     & \multicolumn{2}{c}{RnR+UF} \\
     
     & \textbf{G4o} & \textbf{L3.2}
     & \textbf{G4o} & \textbf{L3.2}
     & \textbf{G4o} & \textbf{L3.2} \\
    \hline

    accuracy & 0.682 & 0.695 & 0.682 & 0.695 & 0.696 & 0.699 \\
    precision & 0.476 & 0.514 & 0.476 & 0.514 & 0.486 & 0.515 \\
    F1 score & 0.617 & 0.548 & 0.617 & 0.548 & 0.611 & 0.533 \\
    TPR/recall & 0.877 & 0.586 & 0.877 & 0.586 & 0.822 & 0.552 \\
    FPR & 0.399 & 0.254 & 0.399 & 0.254 & 0.356 & 0.235 \\
    TNR & 0.601 & 0.746 & 0.601 & 0.746 & 0.644 & 0.765 \\
    FNR & 0.123 & 0.414 & 0.123 & 0.414 & 0.178 & 0.448 \\ \hline
    EHR & 0.351 & 0.129 & .0536 & .0484 & 0.107 & .0492 \\
    EHRC & 0.4 & 0.220 & .0612 & .0865 & 0.129 & .0891 \\
    EPR & 0.318 & 0.221 & 0.422 & 0.287 & 0.384 & 0.285 \\
    EPRC & 0.363 & 0.377 & 0.483 & 0.514 & 0.466 & 0.517 \\
    ERR & 0.576 & 0.320 &  0.378 & 0.253 & 0.408 & 0.249 \\
    ERRC & 0.657 & 0.546 & 0.432 & 0.452 & 0.496 & 0.450 \\
    AECR & .0911 & .0764 & .0296 & .0245 & .0362 & .0238 \\
    avg\#sen, pos & 7.87 & 9.67 & 2.52 & 3.31 & 3.37 & 3.23 \\
    avg\#sen, neg & 6.41 & 6.85 & 2.26 & 2.67 & 2.90 & 2.84 \\
    avg\#sen, all & 7.11 & 8.30 & 2.39 & 3.02 & 3.13 & 3.04 \\
    avg\#retries, pos & 3.61 & 3.37 & 4.61 & 4.37 & 4.61 & 4.37 \\
    avg\#retries, neg & 2.88 & 2.55 & 3.88 & 3.55 & 3.88 & 3.55 \\
    avg\#retries, all & 3.23 & 2.97 & 4.23 & 3.97 & 4.23 & 3.97 \\
    \hline
    \end{tabular}
    \caption{\textbf{ContraDocPaired ablation.} G4o and L3.2 are shorthands for GPT-4o and LLaMA3.2-90B respectively. `pos' refers to datapoints $i$ such that $y_i=\text{Yes}$, `neg' refers to datapoints $i$ such that $y_i=\text{No}$, and `all' refers to all datapoints.}
    \label{tab:ContraDocPaired_ablation}
\end{table*}

\begin{table*}[t]
    \centering
    \begin{tabular}{lcccccccc}
    \hline
     & \multicolumn{2}{c}{DP}
     & \multicolumn{2}{c}{S-con (temp 0.5)}
     & \multicolumn{2}{c}{S-cri}
     & \multicolumn{2}{c}{RnR+CF} \\
     
     & \textbf{G4o} & \textbf{L3.2}
     & \textbf{G4o} & \textbf{L3.2}
     & \textbf{G4o} & \textbf{L3.2}
     & \textbf{G4o} & \textbf{L3.2} \\
    \hline

    accuracy & 0.682 & 0.695 & 0.758 & 0.644 & 0.817 & 0.677 & 0.682 & 0.695 \\
    precision & 0.476 & 0.514 & 0.575 & 0.451 & 0.700 & 0.457 & 0.476 & 0.514 \\
    F1 score & 0.617 & 0.548 & 0.695 & 0.517 & 0.715 & 0.301 & 0.617 & 0.548 \\
    TPR/recall & 0.877 & 0.586 & 0.878 & 0.606 & 0.732 & 0.224 & 0.877 & 0.586 \\
    FPR & 0.399 & 0.254 & 0.297 & 0.339 & 0.144 & 0.119 & 0.399 & 0.254 \\
    TNR & 0.601 & 0.746 & 0.703 & 0.661 & 0.856 & 0.881 & 0.601 & 0.746 \\
    FNR & 0.123 & 0.414 & 0.122 & 0.394 & 0.268 & 0.776 & 0.123 & 0.414 \\ \hline
    EHR & 0 & .00538 & 0.0457 & 0.0455 & 0 & .00521 & .0536 & .0484 \\
    EHRC & 0 & .00917 & 0.0520 & 0.075 & 0 & .0233 & .0612 & .0865 \\
    EPR & 0.400 & 0.266 & 0.360 & 0.228 & 0.371 & 0.118 & 0.422 & 0.287 \\
    EPRC & 0.456 & 0.454 & 0.410 & 0.375 & 0.507 & 0.525 & 0.483 & 0.514 \\
    ERR & 0.351 & 0.223 & 0.393 & 0.265 & 0.317 & 0.102 &  0.378 & 0.253 \\
    ERRC & 0.4 & 0.381 & 0.448 & 0.438 & 0.433 & 0.453 & 0.432 & 0.452 \\
    avg\#sen, pos & 2.04 & 2.53 & 3.17 & 4.61 & 2.02 & 2.44 & 2.52 & 3.31 \\
    avg\#sen, neg & 2.22 & 2.61 & 3.64 & 4.85 & 2.15 & 2.41 & 2.26 & 2.67 \\
    avg\#sen, all & 2.13 & 2.57 & 3.37 & 4.74 & 2.06 & 2.43 & 2.39 & 3.02 \\
    avg\#retries, pos & NA & NA & NA & NA & NA & NA & 4.61 & 4.37 \\
    avg\#retries, neg & NA & NA & NA & NA & NA & NA & 3.88 & 3.55 \\
    avg\#retries, all & NA & NA & NA & NA & NA & NA & 4.23 & 3.97 \\
    \hline
    \end{tabular}
    \caption{\textbf{ContraDocPaired.} G4o and L3.2 are shorthands for GPT-4o and LLaMA3.2-90B respectively. `pos' refers to datapoints $i$ such that $y_i=\text{Yes}$, `neg' refers to datapoints $i$ such that $y_i=\text{No}$, and `all' refers to all datapoints.}
    \label{tab:ContraDocPaired_diff_methods}
\end{table*}


\section{Prompts} \label{app:prompts}

The prompts used are stated here. Note that (i) for redact-and-retry, the same prompt for direct prompting is used during every retry step, and (ii) for self-consistency, the same prompt for direct prompting is used during every sampling step.

\paragraph{Prompt for direct prompting, self-consistency, and redact-and-retry:}

\begin{Verbatim}[formatcom=\color{red},breaklines,fontsize=\tiny,frame=single]
The task is to determine whether the document contains any self-contradictions. If yes, provide evidence by quoting mutually contradictory sentences in a list of strings in Python. If no, then give an empty list. Your response must follow this JSON format (OR options are provided), and provide absolutely nothing else. Strictly follow the double quotation marks and only use single quotations within each sentence.

### JSON format
{
    "judgement": "yes" OR "no",
    "evidence": ["sentence1", "sentence2", ..., "sentenceN"] OR []
}

### document
{document}
\end{Verbatim}

\paragraph{Prompts for filtering (unconstrained):}

\begin{Verbatim}[formatcom=\color{red},breaklines,fontsize=\tiny,frame=single]
You will be given a list of sentences that are flagged to be potentially inconsistent. Your task is to identify all inconsistent sentences and output them in the following JSON format, and provide absolutely nothing else. Strictly follow the double quotation marks and only use single quotations within each sentence.

### JSON format
{
    "evidence": ["sentence1", "sentence2", ..., "sentenceN"] OR []
}

### potentially inconsistent list of sentences
{list of inconsistent sentences}
\end{Verbatim}

\paragraph{Prompts for filtering (constrained):}

\begin{Verbatim}[formatcom=\color{red},breaklines,fontsize=\tiny,frame=single]
You will be given a list of sentences that are flagged to be potentially inconsistent. Your task is to identify all inconsistent sentences and output them in the following JSON format, and provide absolutely nothing else. You must output at least 1 sentence. Strictly follow the double quotation marks and only use single quotations within each sentence.

### JSON format
{
    "evidence": ["sentence1", "sentence2", ..., "sentenceN"]
}

### potentially inconsistent list of sentences
{list of inconsistent sentences}
\end{Verbatim}

\paragraph{Prompts for self-criticism:}

The first prompt to get the answer is the same as the prompt for direct prompting stated above. The second prompt to get the self-criticism is as follows:

\begin{Verbatim}[formatcom=\color{red},breaklines,fontsize=\tiny,frame=single]
Review your previous answer to the task below using the document. If you are very confident about your answer, maintain your answer. Otherwise, update your answer. Present your final answer in the following JSON format (OR options are provided), and provide absolutely nothing else. Strictly follow the double quotation marks and only use single quotations within each sentence.

### JSON format
{
    "judgement": "yes" OR "no",
    "evidence": ["sentence1", "sentence2", ..., "sentenceN"] OR []
}

### Task
The task is to determine whether the document contains any self-contradictions. If yes, provide evidence by quoting mutually contradictory sentences in a list of strings in Python. If no, then give an empty list.

### Previous answer
{previous answer}

### document
{document}
\end{Verbatim}

\section{ContraDoc Details} \label{app:contradoc}

The ContraDoc dataset from \citet{Li2024b} is a human-annotated dataset consisting of self-contradictory documents across varying document domains and lengths and self-contradiction types. More specifically, each positive datapoint in their dataset contains exactly one sentence with one of the following 8 types of self-contradiction:
\begin{enumerate}
    \item \textbf{Negation.} There exists one sentence which is a negation of another sentence. Example: `Zully donated her kidney.' vs.~`Zully never donated her kidney.'
    \item \textbf{Numeric.} There exists a numerical mismatch between sentences. Example: `All the donors are between 20 to 45 years old.' vs.~`Lisa, who donates her kidney, she is 70 years old.'
    \item \textbf{Content.} There exists one sentence changing one or multiple attributes of an event or entity that was previously stated in another sentence. Example: `Zully Broussard donated her kidney to a stranger.' vs.~`Zully Broussard donated her kidney to her close friend.'
    \item \textbf{Perspective/View/Opinion.} Inconsistency in perspective/view/opinion between sentences. Example: `The doctor spoke highly of the project and called it a breakthrough' vs.~`The doctor disliked the project, saying it had no impact at all.'
    \item \textbf{Emotion/Mood/Feeling.} Inconsistency in emotion/mood/feeling between sentences. Example: `The rescue team searched for the boy worriedly.' vs.~`The rescue team searched for the boy happily.'
    \item \textbf{Relation.} Presence of two mutually exclusive relations between entities. Example: `Jane and Tom are a married couple.' vs.~`Jane is Tom’s sister.'
    \item \textbf{Factual.} There exist sentence(s) in disagreement with external world knowledge/facts. Example: `The road T51 was located in New York.' vs.~`The road T51 was located in California.'
    \item \textbf{Causal.} There exist sentences where the effect does not match the cause. Example: `I slam the door.' vs.~`After I do that, the door opens.'
\end{enumerate} 

\end{document}